\documentclass[preprint,12pt]{elsarticle}

\usepackage{amssymb}

\usepackage{graphicx}
\usepackage{subcaption}
\usepackage{amsmath}
\usepackage{amssymb}
\usepackage{mathtools}
\usepackage{amsthm}
\usepackage{amsfonts,bm}
\usepackage{algorithm}
\usepackage{algorithmic}
\usepackage{multirow}
\usepackage{hyperref}
\usepackage[table]{xcolor}
\usepackage{setspace}
\journal{Pattern Recognition}

\begin{document}

\begin{frontmatter}

\title{Molecular Graph Contrastive Learning with Line Graph}

\author[add1]{Xueyuan Chen\corref{equal_contributaion}}
\author[add2]{Shangzhe Li\corref{equal_contributaion}}
\author[add1]{Ruomei Liu}
\author[add3]{Bowen Shi}
\author[add1]{Jiaheng Liu}
\author[add1]{Junran Wu\corref{mycorrespondingauthor}}
\author[add1]{Ke Xu}

\cortext[equal_contributaion]{Equal Contribution.}
\cortext[mycorrespondingauthor]{Corresponding author: wu\_junran@buaa.edu.cn (Junran Wu).}

\affiliation[add1]{organization={State Key Laboratory of Complex $\&$ Critical Software Environment, Beihang University},
            % addressline={37 Xueyuan Road, Haidian District}, 
            city={Beijing},
            % postcode={100191}, 
            % state={Beijing},
            country={China}}

\affiliation[add2]{organization={School of Statistics and Mathematics, Central University of Finance and Economics},
            % addressline={39 South College Road, Haidian District}, 
            city={Beijing},
            % postcode={100081}, 
            % state={Beijing},
            country={China}}

\affiliation[add3]{organization={School of Journalism, Communication University of China},
            % addressline={No.1 Dingfuzhuang East Street}, 
            city={Beijing},
            % postcode={100024}, 
            % state={Beijing},
            country={China}}

\begin{abstract}
Trapped by the label scarcity in molecular property prediction and drug design, graph contrastive learning (GCL) came forward. Leading contrastive learning works show two kinds of view generators, that is, random or learnable data corruption and domain knowledge incorporation. While effective, the two ways also lead to molecular semantics altering and limited generalization capability, respectively. To this end, we relate the \textbf{L}in\textbf{E} graph with \textbf{MO}lecular graph co\textbf{N}trastive learning and propose a novel method termed \textit{LEMON}. Specifically, by contrasting the given graph with the corresponding line graph, the graph encoder can freely encode the molecular semantics without omission. Furthermore, we present a new patch with edge attribute fusion and two local contrastive losses enhance information transmission and tackle hard negative samples. Compared with state-of-the-art (SOTA) methods for view generation, superior performance on molecular property prediction suggests the effectiveness of our proposed framework.
\end{abstract}


\end{frontmatter}

\section{Introduction}
\label{sec:intro}
A deep understanding of molecular properties plays a vital role in the chemical and pharmaceutical domains.
In order to computationally discover novel materials and drugs, the molecules will be abstractly regarded as graphs, in which atoms are vertices and bonds are edges~\cite{ye2022molecular}.
Thus, the marriage between molecular property prediction and graph learning captured a bunch of researchers and showed their happiness in several fields~\cite{ye2022molecular}.
However, this relationship faces the challenges of label scarcity, as deep learning methods are known to consume massive amounts of labeled data, and annotated data is often of limited size and hard to acquire when considering many specific domains. 
In addition, given the immense differentiation among chemical molecules, existing supervised models could be barely reused in unseen cases~\cite{hu2020strategies,rong2020self}.
Therefore, there are increasing demands for molecular representation learning in an unsupervised manner.

Plenty of works have attempted to learn molecule representations discarding the supervision of labels, like masked attribute prediction \cite{hu2020strategies}, graph-level motif prediction \cite{rong2020self}, and graph context prediction \cite{liu2019n}.
In light of the contrastive learning from computer vision, researchers go one step further to model molecules in a contrastive manner with data augmentations \cite{you2020graph,luo2023dual}.
Considering the inherent characteristics of chemical molecules, GCL incorporating well-designed domain knowledge has also shown excellent capacity in molecular properties prediction \cite{sun2021mocl,fang2022molecular}.

Analogously, everything comes with a price. 
Inspecting the generated views in previous molecular contrastive learning unveils two intrinsic limitations.
First, data augmentation-based methods adopting random or learnable corruption (e.g., node/edge dropping and graph generation) would lead to inevitable variance in the crucial semantics and further misguide the contrastive learning \cite{you2020graph,sun2021mocl}.
Second, based on predefined sub-structure substitution rules \cite{sun2021mocl} or contrasted with 3-dimension geometric views \cite{stark20223d}, domain knowledge-based methods intend to alleviate the problem of semantic alteration. While effective, they are stinted to the profound domain knowledge that is unfriendly to researchers without such knowledge, thus limiting their generalization capability in other domains \cite{li2022let}.

\begin{figure}[!tp]
  \centering
  \includegraphics[width=1.\linewidth]{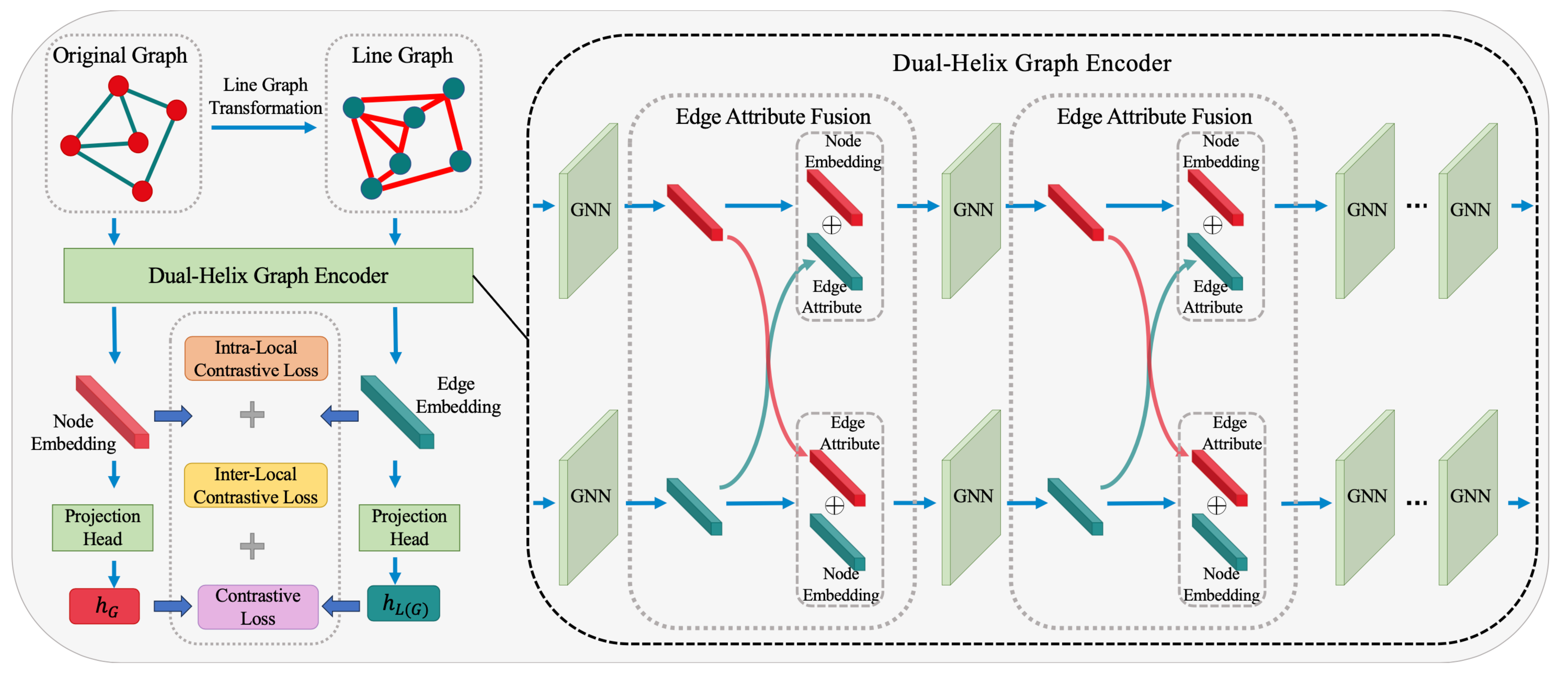}
  \caption{\textbf{Framework overview of LEMON.} Contrasted views consist of the original graph and the corresponding line graph. Input graphs are encoded by a dual-helix graph encoder with edge attribute fusion for information consistency. The whole model is jointly optimized via minimizing the NT-Xent loss and the two local contrastive losses.}
  \label{fig:framework}
\end{figure}

In this context, we are seeking for a decent view that will not be bothered by prefabricated domain knowledge and can maintain the molecular semantic information integrally.
Fortunately, we met the line graph, also known as congruent graph in graph theory \cite{whitney1932congruent,harary1960some}. 
In a line graph, the nodes correspond to the edges of the original graph, and the edges refer to the common nodes of the pair edges in the original graph~\cite{gharaee2021graph}.
In particular, the isomorphism of two line graphs is judged to be consistent with the corresponding two original graphs \cite{whitney1932congruent}, which ensures the congruent semantic structure after line graph transformation.
In light of the line graph, we propose a method termed LEMON to tackle our expectations.

The framework of LEMON is shown in Figure~\ref{fig:framework}. 
Specifically, to fill the framework demanding two views,
all input molecular graphs are transformed into the corresponding line graph. 
On such a basis, LEMON equips with a dual-helix graph encoder to learn the hidden representation of two views.
Note that, due to the different pace of message passing in the line graph and the corresponding original graph, two issues derive from the learning process, that is, information inconsistency and over-smoothing.
For information consistency, we further update the graph encoder with edge attribute fusion to bridge the edge attributes between the two kinds of graphs.
Over-smoothing is addressed by a novel intra-local contrastive loss based on the idea of NT-Xent loss.
Moreover, considering the hard negative samples that have different structural properties but output similar graph embedding, we develop a novel inter-local contrastive loss to address them.

The effectiveness of LEMON is verified under the ubiquitous setting of transfer learning for molecular property prediction \cite{hu2020strategies}. 
Through pre-training on two million molecular graphs from ZINC15, LEMON shows superior performance on all eight benchmarks for molecular property prediction and acquires the highest position on both average ROC-AUC and average ranking.
We delve deeper into the proposed components via analytical experiments to further assess their benefits.
The contributions are elaborated on below:
\begin{itemize}
    \item To the best of our knowledge, we are the first to figure out a way to freely and fully excavate molecular semantics within GCL.
    \item Inspired by the line graph, we present an approach, termed LEMON, to tackle our expectations, in which edge attribute fusion and an intra-local contrastive losses are united to address the concomitant issues.
    \item Considering the correspondence of node embeddings in positive pair samples, we further enhance LEMON with the ability to handle hard negative samples by an inter-local contrastive loss.
    \item Leveraging eight benchmarks for molecular property prediction of transfer learning, LEMON shows its superiority against the SOTA methods.
\end{itemize}

\section{Related Work}
In this section, we elaborate on molecular contrastive view generation, especially the case that is free from intricate domain knowledge.

\subsection{Molecular Contrastive Learning}
Along with the development of GCL, plenty of research efforts have been devoted to designing contrastive learning models for molecular graphs, such MoCL~\cite{sun2021mocl}, KCL~\cite{fang2022molecular} and GeomGCL~\cite{li2022geomgcl}.
Besides random or learnable corruption, several works presented various contrastive learning models to embed the molecular geometry information by means of contrasting the generic 2D graph with its 3D conformers \cite{stark20223d,li2022geomgcl}.
They indeed get rid of the semantic altering issue caused by random corruption on molecular graphs, while introducing another semantic altering issue caused by 3D conformers, because a single 2D molecular graph generally has multiple conformers with different chemical properties~\cite{stark20223d}.
To enhance the performance in molecular property prediction, the domain knowledge-driven contrastive learning frameworks were proposed to preserve the semantics of graphs in the augmentation process \cite{sun2021mocl, fang2022molecular}.
However, their learning capability heavily relies on the dissolved domain knowledge, that is the well-designed substitution rules in MoCL \cite{sun2021mocl} and the prefabricated associations among chemical elements in KCL \cite{fang2022molecular}. Furthermore, the domain knowledge varies across domains, which limits the application of these methods.

\subsection{Line Graph}
The line graph is a classic concept and has a long history in graph theory \cite{whitney1932congruent,harary1960some}. In a line graph, the nodes correspond to the edges of the original graph, and the edges refer to the common nodes of the pair edges in the original graph.
Thus, the graph neural networks (GNNs) built on line graphs are capable of encoding edge features and enhancing feature learning on graphs.
Recently, based on the line graph structures, several line graph neural networks have shown promising performance on various graph-related tasks \cite{chen2018supervised,zhang2023line}.
In the chemistry domain, the structure of a compound can be treated as a graph, where the edges derived from chemical bonds link the corresponding atom nodes.
Thus, the edges in such graphs have different properties and various functions.
In generic GNNs, however, message-passing operations among nodes do not pay enough attention to the edge properties.
Fortunately, the line graph structure enables generic GNNs to take advantage of the edges as equals to nodes \cite{jiang2019censnet,chen2018supervised}. 

This far, little attention has been paid to encoding the molecular semantics integrally without well-designed domain knowledge. In this paper, we revisit the line graph from the angle of GCL. Based on it, we design a novel contrastive model to freely and fully excavate molecular semantics.

\section{Preliminaries}
Before the elaboration of LEMON, here, some preliminary concepts and notations are given.
Let $\mathbb{G} = \{G_1, G_2,\cdots, G_N\}$ be a graph dataset with size $N$, and a molecular graph can be formulated as $G=(V, E, X_V, X_E)$, where $E$ is the edges, $V$ is the vertices, $X_V\in\mathbb{R}^{|V|\times \mathbb{V}}$ denotes node features, and $X_E\in\mathbb{R}^{|E|\times \mathbb{E}}$ stands for edge attributes.

\subsection{Graph Representation Learning}
In generic GNNs, the message-passing scheme is adopted for information transmission among nodes \cite{xu2019powerful,wu2022simple}. 
Through stacking $C$ layers, a GNN will produce a hidden representation $h_v\in\mathbb{R}$ with $C$-hop neighbor information for each node and a feature vector $h_G\in\mathbb{R}$ via a global readout function for the entire graph $G$. Each node $v$ is initialized with node feature $X_v$ and sent to the GNN input.
Formally, the hidden vector of node $v$ in layer $c$ is:
\begin{align}
\hat{h}_v^{(c)} &= \text{AGGREGATE}^{(c)}(\{h_u^{(c-1)}|u\in N(v)\}), \\
h_v^{(c)} &= \text{COMBINE}^{(c)}(\hat{h}_v^{(c)}, h_v^{(c-1)}),
\end{align}
where $h_v^{(c)}$ represents the output vector after $c$ iterations of node $v$, $N(\cdot)$ covers the 1-hop neighbors, AGGREGATE denotes the crucial message-passing scheme in GNNs, and COMBINE is used to update the hidden feature of $v$ via merging information from its neighbors and itself.
Finally, a GNN can produce the feature vector $h_G$ of the entire graph with a prefabricated readout function:
\begin{equation}
h_G = \text{READOUT}(\{h_v|v\in\mathcal{V}\}),
\end{equation}
where READOUT aggregates the final set of node representations and is generally a summation or averaging function.

\subsection{Graph Contrastive Learning}
In a generic GCL model, two correlated views from the same graph $G_i$ are required for contrasting and generally produced by two augmentation operations. Here, the two views for contrast can be termed as $\tilde{G}^1_i$ and $\tilde{G}^2_i$. Then, a graph encoder and a projection head are stacked behind the two augmentation operators to map the correlated views into an embedding space and yield corresponding feature vectors $h_i^1$ and $h_i^2$.
The released hidden representations are supposed to contain the essential information underlying the raw graph so that they are capable of recognizing themselves from the others. Thus, the objective of GCL is to maximize the consensus between the two positive views via the widespread NT-Xent loss \cite{chen2020simple}:
\begin{equation}
\mathcal{L}_i = -\log\frac{e^{sim(h_i^1, h_i^2)/\tau}}{\sum^N_{j=1,j\neq i}e^{sim(h_i^1, h_j^2)/\tau}},
\end{equation}
where $N$ is the batch size, $\tau$ refers to the temperature parameter, and $sim(h^1, h^2)$ denotes a cosine similarity function $\frac{h^{1\top} h^2}{||h^1||\cdot||h^2||}$.
The numerator part is the similarity of the correlated views as positive pair. The rest pairs that consist of views from different graphs are regarded as negative pairs and act as the denominator part.
Note that the negative pairs can come from two directions, put differently, $h_i^1$ can pair with all $h_j^2$, and $h_i^2$ can pair with all $h_j^1$. In this work, the loss function is also comprised of two directions.

\section{Methodology}
In this section, we bring about the proposed GCL framework, termed \textit{LEMON}, by revisiting the line graph of corresponding molecules. Given the issue of label scarcity in real-world graph data, LEMON is designed to encode the molecular semantics integrally and free from well-designed domain knowledge.
Specifically, to produce two contrastive views without any loss of molecular semantics, we first need to transform the given molecule to the corresponding line graph. On such a basis, LEMON equips with a dual-helix graph encoder to learn the hidden representation of two views with edge attribute fusion.
In particular, besides the ubiquitous contrastive loss for the readout graph representations, we further propose two local contrastive losses to tackle hard negative samples and alleviate the over-smoothing 
issue in deep GNNs.
Next, we elaborate on the LEMON framework below.

\begin{figure}[!t]
  \centering
  \begin{subfigure}{.2\textwidth}
    \centering
    \includegraphics[width=\linewidth]{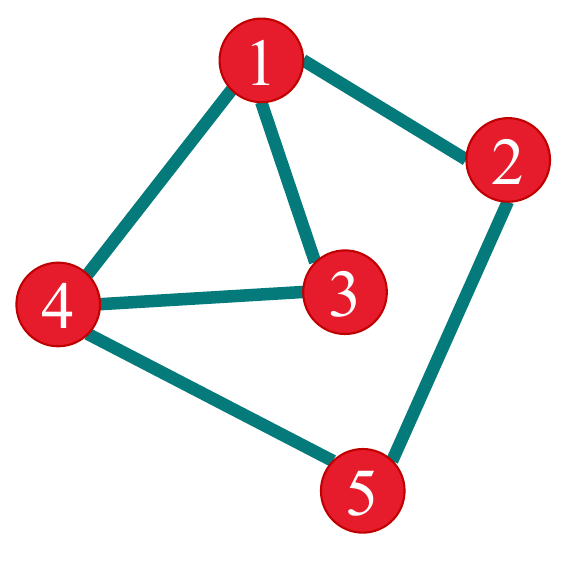}
    % \caption{Graph $G$}
  \end{subfigure}
  \begin{subfigure}{.2\textwidth}
    \centering
    \includegraphics[width=\linewidth]{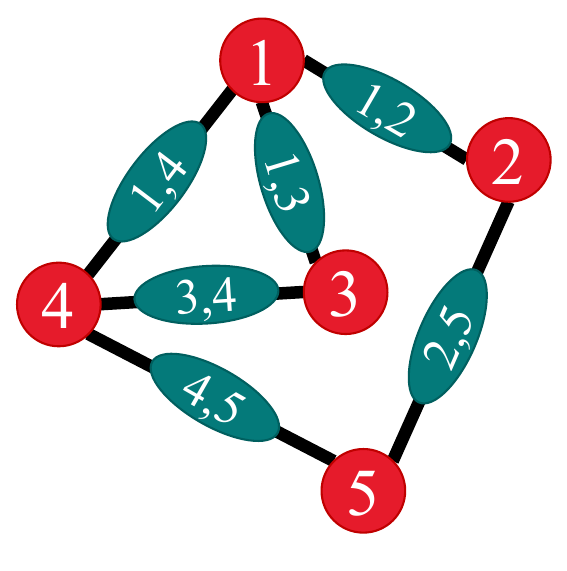}
  \end{subfigure} 
  \begin{subfigure}{.2\textwidth}
    \centering
    \includegraphics[width=\linewidth]{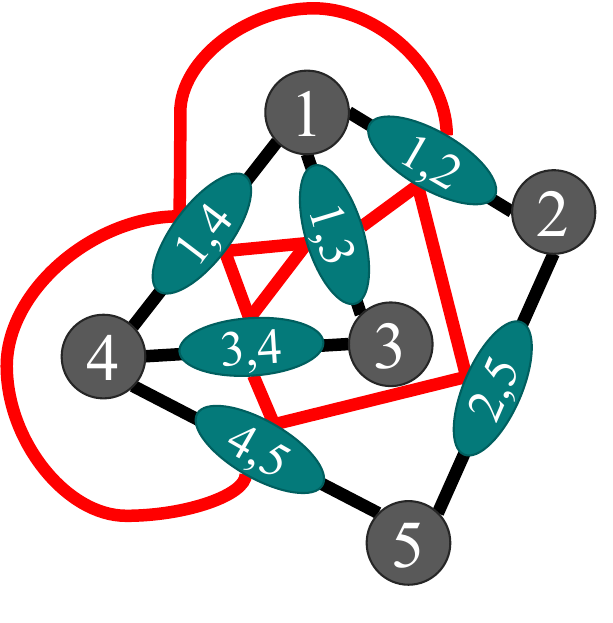}
  \end{subfigure}
  \begin{subfigure}{.2\textwidth}
    \centering
    \includegraphics[width=\linewidth]{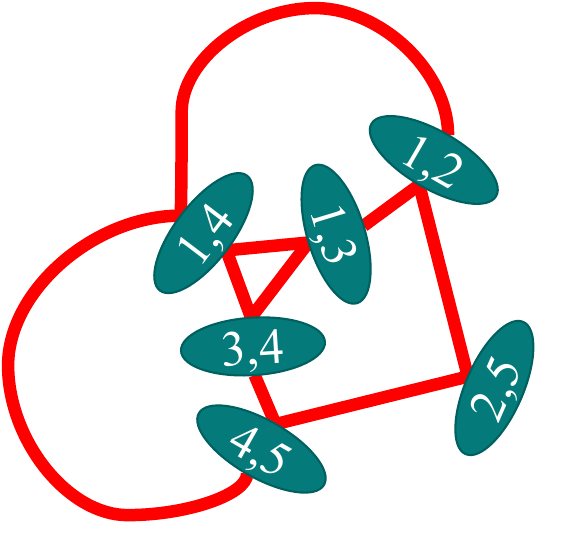}
  \end{subfigure} 
  \\
  \begin{subfigure}{.2\textwidth}
    \centering
    \caption{Graph $G$}
  \end{subfigure}
  \begin{subfigure}{.2\textwidth}
    \centering
    \caption{Vertices of $L(G)$ born from edges of $G$}
  \end{subfigure} 
  \begin{subfigure}{.25\textwidth}
    \centering
    \caption{Add edges in $L(G)$}
  \end{subfigure}
  \begin{subfigure}{.2\textwidth}
    \centering
    \caption{Line graph $L(G)$}
  \end{subfigure} 
  \caption{\textbf{An illustration of line graph transformation.} (a) shows a simple undirected graph $G$; (b) reveals the derivation of vertices in line graph, every vertex of line graph is marked with green and labeled with the pair nodes of the corresponding edge in $G$; (c) establishes the associations in $L(G)$ according to the common nodes owned by two edges; (d) delivers the output line graph $L(G)$ after transformation.} 
  \label{fig:lg_transformation}
\end{figure}

\subsection{Line Graph Transformation}
\label{sec:lg_trans}
An illustration of line graph transformation from a simple graph is first shown in Figure~\ref{fig:lg_transformation}. 
The output line graph $L(G)$ after transformation from the input simple undirected graph $G=(V, E)$ is such a graph that reveals the adjacencies of edges in $G$. 
Specifically, each node of $L(G)$ denotes an edge in $G$ and each edge in $L(G)$ indicates that the corresponding two vertices have a common node in $G$. 
Formally, the line graph can be written as $L(G)=(V_L, E_L)$, where $V_L=\{(v_i, v_j)|(v_i, v_j)\in E)\}$ and $E_L=\{((v_i, v_j), (v_j, v_k))|\{(v_i,v_j), (v_j,v_k)\}\subset E\}$.
At this point, we have settled the topology transformation of line graphs. Besides the relationships among nodes and edges, the node and edge attributes underlying the molecular graph should also be delivered to the corresponding line graph. 
In this paper, based on the one-to-one correspondence between the vertices of $L(G)$ and the edges of $G$, the node attributes of $L(G)$ can be directly obtained from the edge attributes of original graphs, that is, $X_{V_L}=X_E$. As for the edge attributes of $L(G)$, because several edges in the line graph would correspond to the same node in the original graph, a mapping function with such relationships is required to endow the line graph edge with the original node attribute. 
In the light of $E_L$ after the line graph transformation, the edge attribute mapping function can be formulated as $M(e_L) = (v_i, v_j) \cap (v_j, v_k)$, thus the edge attributes of the line graph can be obtained via $X_{E_L} = MX_V$. Finally, the line graph of a molecular graph is given by $L(G)=(V_L, E_L, X_{V_L}, X_{E_L})$. 
According to Roussopoulos’s algorithm~\cite{roussopoulos1973max}, the time complexity of line graph transformation is $O(\max(|V|, |E|))$.

As stated in the Whitney graph isomorphism theorem \cite{whitney1932congruent}, the isomorphism of two line graphs is judged to be consistent with the corresponding two original graphs, which convinces us that the semantic structure information of $G$ is encoded in its line graph. 
In particular, considering the transformation process of line graph, a vertex with $e$ edges in $G$ will produce $e\times(e-1)/2$ edges in $L(G)$. Meanwhile, the message-passing frequency around this node will drift from $O(e)$ in $G$ to $O(e^2)$ in $L(G)$, put differently, this node feature in $G$ is only passed to $e$ neighbors, while the corresponding line graph will pass such information to $e\times(e-1)/2$ nodes.
Actually, because the line graph is static, the time required by LEMON for pre-training is much less than the time needed by models with data augmentation.
While this nature of the line graph could cause two inevitable issues in the contrastive learning framework with stacked graph convolutional layers, that is information inconsistency and over-smoothing.
In this paper, we propose two approaches, edge attribute fusion and an intra-local contrastive loss, to alleviate the two issues and strengthen molecular representation learning.

\subsection{Edge Attribute Fusion}
In the chemistry domain, the structure of a compound can be treated as a graph, where the edges derived from the chemical bonds link the corresponding atom nodes.
Thus, the edges in such graphs have different properties and various functions.
Besides the topology information weaved by atoms, a well-designed graph convolution with edge attributes plays a crucial role in molecular property and protein function prediction.

Given a molecule, its input node features and edge features are both represented as a 2-dimensional categorical vector, denoted as $X_V\in\mathbb{R}^{|V|\times 2}$ and $X_E\in\mathbb{R}^{|E|\times 2}$, respectively. 
In previous works regarding molecular property prediction \cite{hu2020strategies}, the raw node categorical vectors are embedded in the input layer by
\begin{equation}
h_v^{(0)} = \text{EMBEDDING}(x_v^0) + \text{EMBEDDING}(x_v^1),
\end{equation}
where $x_v^0$ and $x_v^1$ are the atomic number and chirality tag of node $v$, respectively. $\text{EMBEDDING}(\cdot)$ denotes an embedding function that transfers a single integer into a $d$-dimensional vector space. Meanwhile, the raw edge categorical vectors are embedded in each layer by
\begin{equation}
h_e^{(c)} = \text{EMBEDDING}(x_e^0) + \text{EMBEDDING}(x_e^1),
\end{equation}
where $x_e^0$ and $x_e^1$ represent the bond type and bond direction, respectively, and $c$ denotes the depth of graph encoder. 
After $c$ iterations, the node representation can be updated by
\begin{equation}
h_v^{(c)} = \sigma(\text{MLP}^{(c)}( h_v^{(c-1)}+ \sum_{u\in N(v)}h_u^{(c-1)} + \sum_{e\in\{(v,u)|u\in N(v)\cup\{v\}\}}h_e^{(c-1)})),
\end{equation}
where $\sigma(\cdot)$ is an activation function, and $(v, v)$ represents the self-loop edge.

Under this GNN architecture, the output molecular representations will be decorated with edge attributes.
However, as discussed above, there is a significant difference in message-passing frequency between the original graph and the corresponding line graph, which could lead to information inconsistency between the outputs. 
Here, we present a novel \textit{edge attribute fusion} approach to tackle this issue. Specifically, we bridge the edge information between the molecular graph and line graph to help the original graph encoder keep pace with the line graph encoder. 
The edge and node embeddings are still employed as the initial edge attributes in the first layer (i.e., $c=0$). As for $c\geq 1$, the edge attributes of the original graph are obtained from the node hidden vectors in $L(G)$, which is formally given by
\begin{equation}
h_{G\cdot (v_i, v_j)}^{(c)} = h_{L(G)\cdot (v_i, v_j)}^{(c-1)},
\end{equation} 
where $(v_i, v_j)\in E$ and $(v_i, v_j) \in V_L$.
Correspondingly, the edge attributes of the line graph can be derived from the node hidden features in the original graph, which is formally formulated as:
\begin{equation}
h_{L(G)\cdot ((v_i, v_j),(v_j, v_k))}^{(c)} = h_{G\cdot v_j}^{(c-1)},
\end{equation}
where $(v_i, v_j)\in E$, $ (v_j, v_k) \in E$ and $((v_i, v_j),(v_j, v_k))\in E_L$.
Based on the dual-helix graph encoder with edge attribute fusion, the hidden features from the line graph will be dissolved into the original graph representations, allowing information consistency between the two contrastive views and enhancing molecular representation learning.

\subsection{Intra-Local Contrastive Loss}
In this part, we look forward to tackling the over-smoothing issue introduced by the line graph. Motivated by NT-Xent loss for contrastive learning, an intra-local contrastive loss is proposed. 
The design concept of NT-Xent loss aims to maximize the representation similarities of positive pairs consisting of hidden features of the same molecules and enforce dissimilarity of negative pairs comprising hidden features of different molecules simultaneously.
Similarly, the proposed intra-local contrastive loss seeks to optimize the consensus between the same nodes as opposed to different nodes in a single graph.
Considering the one-to-one correspondence between the edges in $G$ and the vertices in $L(G)$, 
the contrastive samples of this loss are composed of the edge hidden features in $G$ and the node hidden features in $L(G)$. Thus, given graph $G$, the intra-local contrastive loss of one edge pair is
 % formally defined as:
\begin{equation}
\mathcal{L}_{IntraC}^{e_i} = -\log\frac
{e^{sim(\tilde{h}_{G\cdot e_i}, h_{L(G)\cdot e_i})/\tau}}
{\sum^{|E|}_{j=1,j\neq i}e^{sim(\tilde{h}_{G\cdot e_i}, h_{L(G)\cdot e_j})/\tau}},
\end{equation}
where $e_i=(v_m, v_n)$, $e_i\in E$ and $(v_m, v_n)\in V_L$. In particular, the edge representations from $G$ are formed by such hidden features of the two endpoints of each edge, that is, $\tilde{h}_{G\cdot e_i}=\text{MLP}([h_{v_m},h_{v_n}])$.
In light of this contrastive loss designed inside the graph, we look forward to reducing the similarity between different nodes and further alleviating the over-smoothing.

\subsection{Hard Negative Samples}
\begin{figure}[!t]
  \centering
  \begin{subfigure}{0.46\linewidth}
    \centering
    \includegraphics[width=\linewidth]{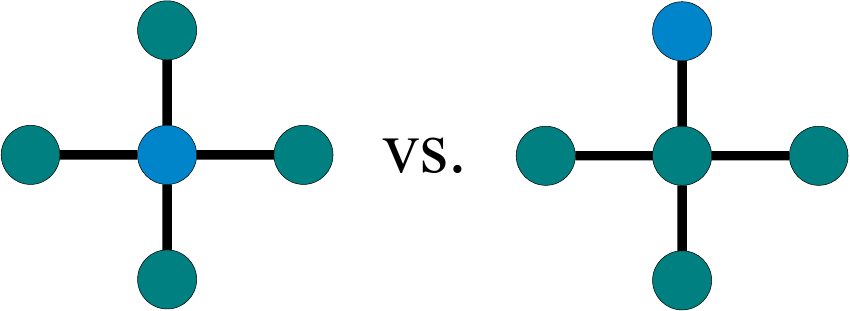}
    \caption{Same structure but different node features.}
    \label{fig:hard_negative_a}
  \end{subfigure}
  \begin{subfigure}{0.46\linewidth}
    \centering
    \includegraphics[width=\linewidth]{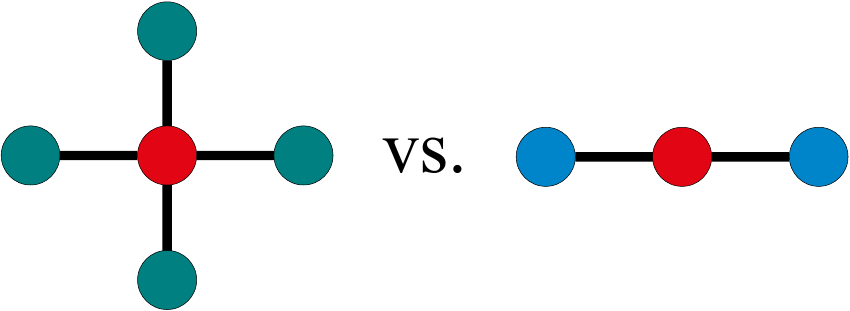}
    \caption{Different structure and node features.}
    \label{fig:hard_negative_b}
  \end{subfigure}
  \caption{\textbf{Illustration of hard negative samples.} Via contrasting the graph embeddings, the pre-trained model is hard to distinguish this two kinds of graphs.} 
  \label{fig:hard_negative}
\end{figure}

Among the research on contrastive learning, hard negative samples are quite ubiquitous and have great potential to improve model performance~\cite{gong2023improving}. However, little attention has been drawn to the hard negative samples within current contrastive learning for molecular pre-training.
As shown in Figure~\ref{fig:hard_negative}, there are two kinds of hard negative samples.
In Figure~\ref{fig:hard_negative_a}, this negative pair has the same topology but different features. Let $h_{color}$ ($g$ for green, $b$ for blue) denote the node features, the graph representations would be similar after pooling as $f_R(4\times h_g + f_b) = f_R(4\times h_g + f_b)$.
In Figure~\ref{fig:hard_negative_b}, this negative pair has different topology and node features.
Let $h_b=2h_g$, the graph representations would be also similar through summation or maximization pooling function.
Given the common design of current contrastive learning for graph classification, the model is encouraged to enlarge the distance of negative pairs via graph representations, which may fail with the two kinds of negative samples in Figure~\ref{fig:hard_negative} and deteriorate model performance.

Despite the hardness of distinguishing the two kinds of negative samples from the graph representations, we may be able to spot some opportunities from the local features.
For example, the green nodes could be effortlessly identified from the blue nodes in Figure~\ref{fig:hard_negative_a}, and the red nodes also have obvious differences compared to the green and blue nodes in Figure~\ref{fig:hard_negative_b}.
Therefore, in order to cultivate the ability of contrastive learning to tackle hard negative samples, we propose a local contrastive loss with the node embeddings.

\begin{algorithm}[!tp]
\small
\caption{Pre-training process of LEMON}
\label{code:pretrain} 
\textbf{Input:} Unlabeled dataset $\mathbb{G}$; graph encoder $f_{GNN}$; projection head $g$. \\
\textbf{Hyper-parameter:} $\alpha$ and $\beta$ for loss weight controlling. \\
\textbf{Output:} pre-trained graph encoder $f_{GNN}$

\begin{algorithmic}[1]
\STATE Initialize graph encoder $f_{GNN}$;
\STATE // \texttt{Line graph transformation};
\STATE Transform dataset $\mathbb{G}$ to corresponding line graph dataset $\mathbb{L}$;
\FOR{\textit{each epoch}} {
  \FOR{\textit{each mini-batch}} {
    \STATE // \texttt{Initialize node features};
    \STATE Obtain the initial node feature $h_v^0$ and $h_e^0$ via Eq.5 and Eq.6.
    \STATE // \texttt{Graph and line graph encoding};
    \FOR{\textit{each layer $f_{GNN}^c$ of graph encoder $f_{GNN}$}} {
      \STATE // \texttt{Edge attribute fusion};
      \STATE For $c=0$, take $h_e^0$ as the edge attribute for original graph and $h_v^0$ as the edge attribute for line graph.
      \STATE For $c>0$, take node embedding $h_{L(G)\cdot (v_i, v_j)}^{(c-1)}$ from the line graph as the edge attribute for the original graph via Eq.8; take the node embedding $h_{G\cdot v_j}^{(c-1)}$ from the original graph as the edge attribute for the line graph via Eq.9;
      \STATE // \texttt{Graph message passing};
      \STATE Obtain the node representation $h_v^c$ for original graph and $h_e^c$ for line graph via Eq.7;
    }\ENDFOR
    \STATE // \texttt{Calculate two local losses};
    \STATE Calculate $\mathcal{L}_{IntraC}$ for mini-batch via Eq.10;
    \STATE Calculate $\mathcal{L}_{InterC}$ for mini-batch via Eq.11;
    \STATE // \texttt{Graph readout};
    \STATE Obtain the graph representation $h_G$ and $h_{L(G)}$ via Eq.3;
    \STATE // \texttt{Calculate graph contrastive loss};
    \STATE Calculate $\mathcal{L}_G$ via Eq.4;
    \STATE // \texttt{Model optimization};
    \STATE Obtain pre-training objective $\mathcal{L} = \mathcal{L}_{G}+\alpha\mathcal{L}_{InterC}+\beta\mathcal{L}_{IntraC}$; 
    \STATE Update encoder $f_{GNN}$ and projection head $g$ to minimize $\mathcal{L}$;
  }\ENDFOR
}\ENDFOR
\STATE return $f_{GNN}$;
\end{algorithmic} 
\end{algorithm}

\noindent \textbf{Inter-local contrastive loss.}
Our critical insight is that current models lack the ability to separate the hard negative samples via graph representations, while a contrastive angle based on node representations would be helpful in this scene.
Specifically, as we are capable of enforcing dissimilarity between different edge representations in the same graph, we move forward to generalizing the edge dissimilarity to all contrasted samples. Formally, the inter-local contrastive loss can be formulated as:
\begin{equation}
\mathcal{L}_{InterC}^{e_i} = -\log\frac
{e^{sim(\tilde{h}_{G\cdot e_i}, h_{L(G)\cdot e_i})/\tau}}
{\sum_{\hat{G}\neq G}^{\mathbb{G}}\sum^{|E_{\hat{G}}|}_{j=1}e^{sim(\tilde{h}_{G\cdot e_i}, h_{L(\hat{G})\cdot e_j})/\tau}},
\end{equation}
where $e_i\in E_{G}$, $e_j \in V_{L(\hat{G})}$ and $\mathbb{G}$ represents a training batch. Note that the negative pairs of the inter-local contrastive loss also come from two directions.

\noindent \textbf{Remark.} Although the inter-local contrastive loss for hard negative samples is relatively simple in its presentation, it is not easy to implement when the two graphs of a positive pair have different structures due to corruption. This limitation may account for why previous models have overlooked this approach and consequently produced sub-optimal performance.

Currently, we have presented the main components of the proposed LEMON that aims to help molecular contrastive learning maintain the semantics and tackle hard negative samples.
For unsupervised molecular graph representation learning, the objective function of LEMON for pre-training is
\begin{equation}
\text{min}\,\,\mathcal{L} = \mathcal{L}_{G}+\alpha\mathcal{L}_{InterC}+\beta\mathcal{L}_{IntraC},
\end{equation}
where $\mathcal{L}_{G}$ denotes the NT-Xent loss, $\alpha$ and $\beta$ are two hyper-parameters for loss weight controlling. The overall pre-training process of LEMON is depicted in Algorithm~\ref{code:pretrain} As can be seen, the pre-training process commences with the initialization of the graph encoder and the transformation of the pre-training graph dataset into a corresponding line graph dataset. 
As for the training process, each iteration initializes node features and performs graph and line graph encoding. This encoding process involves edge attribute fusion and graph message passing, which are executed within each layer of the graph encoder.
While finishing graph encoding, the algorithm calculates two local losses, $\mathcal{L}_{IntraC}$ and $\mathcal{L}_{InterC}$. It then obtains the graph representation via a graph readout function and calculates the graph contrastive loss, $\mathcal{L}_G$. The final pre-training objective, $\mathcal{L}$, is the weighted summation of $\mathcal{L}_{G}$, $\mathcal{L}_{InterC}$, and $\mathcal{L}_{IntraC}$. At the last of each training batch, the graph encoder and projection head are updated to minimize the objective function.

\section{Experiment}
\label{sec:exp}
Now, we are devoted to evaluating LEMON with extensive experiments \footnote{The code of LEMON is available at \url{https://github.com/RyanChen227/LEMON}.}. 
Following the procedure of pre-training and fine-tuning, we validate the effectiveness of LEMON against SOTA competitors for view generation. Furthermore, we carry out analytical studies to assess each proposed component.

\subsection{Experimental Setup}
\label{sec:exp_setup}
To be in line with the previous GCL methods and make the comparisons fair, we follow the experimental setup under the guidance of \citet{hu2020strategies}.

\noindent \textbf{Pre-training dataset.} 
ZINC15 \cite{sterling2015zinc} dataset is adopted for LEMON pre-training. In particular, a subset with two million unlabeled molecular graphs are sampled from the ZINC15.

\noindent \textbf{Pre-training details.} 
In the graph encoder setting in \citet{hu2020strategies}, a Graph Isomorphism Network (GIN) \cite{xu2019powerful} with five convolutional layers is adopted for message passing. In particular, the hidden dimension is fixed to 300 across all layers and a pooling readout function that averages graph nodes is hired for NT-Xent loss calculation with the scale parameter $\tau = 0.1$. The hidden representations at the last layer are injected into the average pooling function.
An Adam optimizer is employed to minimize the integrated losses produced by the 5-layer GIN encoder. The batch size is set as 256, and all training processes will run 100 epochs.
The two hyper-parameters, $\alpha$ and $\beta$, which control the weight of the loss, are tuned within the range of $[0.01, 0.1, 1, 10, 100]$. The optimal combination is determined based on the performance on the validation sets.

\begin{table}[!t]
\centering
\caption{Summary statistics of ubiquitous benchmarks from MoleculeNet.}
\label{tab:data-stat-molecule}
\resizebox{0.85\linewidth}{!}{%
\begin{tabular}{l|c|ccccc}
\hline \hline
Dataset & Utilization & Avg.\#Degree & Avg.\#Node & \#Tasks & \#Graphs \\ \hline \hline
ZINC15  & Pre-Training & 57.72 & 26.63 & & 2,000,000 \\ \hline
BBBP    & Finetuning & 51.90 & 24.06 & 1       & 2,039 \\
BACE    & Finetuning & 73.71 & 34.08 & 1       & 1,513  \\
HIV     & Finetuning & 54.93 & 25.51 & 1       & 41,127 \\
ClinTox & Finetuning & 55.76 & 26.15 & 2       & 1,477 \\
MUV     & Finetuning & 52.55 & 24.23 & 17      & 93,087 \\
Tox21   & Finetuning & 38.58 & 18.57 & 12      & 7,831 \\
SIDER   & Finetuning & 70.71 & 33.64 & 27      & 1,427 \\
ToxCast & Finetuning & 38.52 & 18.78 & 617     & 8,576 \\ \hline \hline
\end{tabular}%
}
\end{table}

\noindent \textbf{Fine-tuning dataset.} 
We employ the eight ubiquitous benchmarks from the MoleculeNet dataset \cite{wu2018moleculenet} to validate LEMON as downstream experiments. These benchmarks include a variety of molecular tasks like physical chemistry, quantum mechanics, physiology, and biophysics. 
For dataset split, the scaffold split scheme \cite{chen2012comparison} is adopted for train/validation/test set generation. 
Table~\ref{tab:data-stat-molecule} summarizes the basic characteristics of the datasets, such as the size, tasks and molecule statistics.

\begin{table}[!tp]
\centering
\caption{Average test ROC-AUC (\%) $\pm$ Std. over different 10 runs of LEMON along with all baselines on eight downstream molecular property prediction benchmarks. The results of baselines are derived from the published works. \textbf{Bold} indicates the best performance among all baselines. Avg. shows the average ROC-AUC over all datasets. A.R. denotes the average rank and - indicates the data missing in such works.}
\label{tab:results}
\resizebox{\textwidth}{!}{%
\begin{tabular}{l|cccccccc|cc}
\hline
Dataset & BBBP & Tox21 & ToxCast & SIDER & ClinTox & MUV & HIV & BACE & Avg. & A.R. \\ \hline
No Pre-Train & 65.8$\pm$4.5 & 74.0$\pm$0.8 & 63.4$\pm$0.6 & 57.3$\pm$1.6 & 58.0$\pm$4.4 & 71.8$\pm$2.5 & 75.3$\pm$1.9 & 70.1$\pm$5.4 & 66.96 & 19.6 \\ \hline
Infomax & 68.8$\pm$0.8 & 75.3$\pm$0.6 & 62.7$\pm$0.4 & 58.4$\pm$0.8 & 69.9$\pm$3.0 & 75.3$\pm$2.5 & 76.0$\pm$0.7 & 75.9$\pm$1.6 & 70.29 & 17.4 \\
EdgePred & 67.3$\pm$2.4 & 76.0$\pm$0.6 & 64.1$\pm$0.6 & 60.4$\pm$0.7 & 64.1$\pm$3.7 & 74.1$\pm$2.1 & 76.3$\pm$1.0 & 79.9$\pm$0.9 & 70.28 & 14.4 \\
AttrMasking & 64.3$\pm$2.8 & 76.7$\pm$0.4 & 64.2$\pm$0.5 & 61.0$\pm$0.7 & 71.8$\pm$4.1 & 74.7$\pm$1.4 & 77.2$\pm$1.1 & 79.3$\pm$1.6 & 69.90 & 13.7 \\
ContextPred & 68.0$\pm$2.0 & 75.7$\pm$0.7 & 63.9$\pm$0.6 & 60.9$\pm$0.6 & 65.9$\pm$3.8 & 75.8$\pm$1.7 & 77.3$\pm$1.0 & 79.6$\pm$1.2 & 70.89 & 13.2 \\
GraphMAE & 72.0$\pm$0.6 & 75.5$\pm$0.6 & 64.1$\pm$0.3 & 60.3$\pm$1.1 & 82.3$\pm$1.2 & 76.3$\pm$2.4 & 77.2$\pm$1.0 & 83.1$\pm$0.9 & 73.85 & 9.8 \\ \hline
GraphCL & 69.68$\pm$0.67 & 73.87$\pm$0.66 & 62.40$\pm$0.57 & 60.53$\pm$0.88 & 75.99$\pm$2.65 & 69.80$\pm$2.66 & 78.47$\pm$1.22 & 75.38$\pm$1.44 & 70.77 & 16.5 \\
JOAO(v2) & 71.39$\pm$0.92 & 74.27$\pm$0.62 & 63.16$\pm$0.45 & 60.49$\pm$0.74 & 80.97$\pm$1.64 & 73.67$\pm$1.00 & 77.51$\pm$1.17 & 75.49$\pm$1.27 & 72.12 & 14.5 \\
LP-Info & 71.40$\pm$0.55 & 74.54$\pm$0.45 & 63.04$\pm$0.30 & 59.70$\pm$0.43 & 74.81$\pm$2.73 & 72.99$\pm$2.28 & 76.96$\pm$1.10 & 80.21$\pm$1.36 & 71.71 & 15.4 \\
AD-GCL & 70.01$\pm$1.07 & 76.54$\pm$0.82 & 63.07$\pm$0.72 & 63.28$\pm$0.79 & 79.78$\pm$3.52 & 72.30$\pm$1.61 & 78.28$\pm$0.97 & 78.51$\pm$0.80 & 72.72 & 11.3 \\
AutoGCL & 73.36$\pm$0.77 & 75.69$\pm$0.29 & 63.47$\pm$0.38 & 62.51$\pm$0.63 & 80.99$\pm$3.38 & 75.83$\pm$1.30 & 78.35$\pm$0.64 & 83.26$\pm$1.13 & 74.18 & 7.4 \\
RGCL & 71.42$\pm$0.66 & 75.20$\pm$0.34 & 63.33$\pm$0.17 & 61.38$\pm$0.61 & 83.38$\pm$0.91 & 76.66$\pm$0.99 & 77.90$\pm$0.80 & 76.03$\pm$0.77 & 73.16 & 10.5 \\
D-SLA & 72.60$\pm$0.79 & 76.81$\pm$0.52 & 64.24$\pm$0.50 & 60.22$\pm$1.13 & 80.17$\pm$1.50 & 76.64$\pm$0.91 & 78.59$\pm$0.44 & 83.81$\pm$1.01 & 74.14 & 6.4 \\
HGCL & 73.60$\pm$1.30 & - & 64.00$\pm$0.40 & - & 84.20$\pm$0.80 & 78.30$\pm$1.20 & 78.80$\pm$0.90 & 80.20$\pm$1.00 & - & 5.1 \\
MACL & 67.98$\pm$0.97 & 74.39$\pm$0.29 & 62.96$\pm$0.28 & 61.46$\pm$0.39 & 78.13$\pm$4.29 & 72.77$\pm$1.25 & 77.56$\pm$1.12 & 76.07$\pm$0.90 & 71.42 & 15.1 \\
GraphACL & 73.30$\pm$0.50 & 76.80$\pm$0.30 & 64.10$\pm$0.40 & 62.60$\pm$0.60 & 85.00$\pm$1.60 & 76.90$\pm$1.20 & 78.90$\pm$0.70 & 80.40$\pm$0.70 & 74.75 & 4.3 \\
SimGRACE & 71.25$\pm$0.86 & - & 63.36$\pm$0.52 & 60.59$\pm$0.96 & - & - & - & - & - & 13.3 \\
GraphLoG & 72.5$\pm$0.8 & 75.7$\pm$0.5 & 63.5$\pm$0.7 & 61.2$\pm$1.1 & 76.7$\pm$3.3 & 76.0$\pm$1.1 & 77.8$\pm$0.8 & 83.5$\pm$1.2 & 73.36 & 9.3 \\
MGSSL & 69.7$\pm$0.9 & 76.5$\pm$0.3 & 64.1$\pm$0.7 & 61.8$\pm$0.8 & 80.7$\pm$2.1 & 78.7$\pm$1.5 & 78.8$\pm$1.2 & 79.1$\pm$0.9 & 73.68 & 7.5 \\
GROVER & 68.0$\pm$1.5 & 76.3$\pm$0.6 & 63.4$\pm$0.6 & 60.7$\pm$0.5 & 76.9$\pm$1.9 & 75.8$\pm$1.7 & 77.8$\pm$1.4 & 79.5$\pm$0.8 & 72.30 & 12.4 \\
MoleBERT & 71.9$\pm$1.6 & 76.8$\pm$0.5 & 64.3$\pm$0.2 & 62.8$\pm$1.1 & 78.9$\pm$3.0 & 78.6$\pm$1.8 & 78.2$\pm$0.8 & 80.8$\pm$1.4 & 74.04 & 5.7 \\ \hline
LEMON & \textbf{73.71$\pm$1.08} & \textbf{77.45$\pm$0.57} & \textbf{65.07$\pm$0.52} & \textbf{64.27$\pm$0.88} & \textbf{85.88$\pm$3.16} & \textbf{79.42$\pm$4.26} & \textbf{79.34$\pm$1.08} & \textbf{87.78$\pm$1.35} & \textbf{76.62} & \textbf{1.0} \\ \hline
\end{tabular}%
}
\end{table}

\noindent \textbf{Fine-tuning details.} 
For downstream tasks, a linear layer is stacked after the pre-trained graph encoders for final property prediction. The downstream model still employs the Adam optimizer for 100 epochs fine-tuning. The learning rate is 0.001 and the batch size is 32.
To mitigate the risk of over-fitting, the dropout ratio for each layer is configured to 0.5.
All experiments on each dataset are performed for ten runs with different seeds, and the results are the averaged ROC-AUC scores (\%) $\pm$ standard deviations.

\noindent \textbf{Baselines.} In this paper, we choose the SOTA competitors that follows the experimental setup in \citet{hu2020strategies}, including EdgePred, AttrMsking, ContexPred \cite{hu2020strategies}, Infomax \cite{velickovic2019deep}, and GraphMAE \cite{hou2022graphmae}, GraphCL~\cite{you2020graph}, JOAO(v2)~\cite{you2021graph}, LP-Info~\cite{you2022bringing}, AutoGCL~\cite{yin2022autogcl}, RGCL~\cite{li2022let} , D-SLA~\cite{kim2022graph}, HGCL~\cite{ju2023unsupervised}, MACL~\cite{huang2023model}, GraphACL~\cite{luo2023self}, SimGRACE~\cite{xia2022simgrace}, GraphLoG~\cite{xu2021self}, MGSSL~\cite{zhang2021motif}, GROVER~\cite{rong2020self}, and MoleBERT~\cite{xia2022mole}.

\subsection{Results}
The results of LEMON along with SOTA competitors for molecular property prediction on eight benchmarks are shown in Table~\ref{tab:results}. 
To summarize, the proposed GCL framework with the line graph, LEMON, obtains superior performance compared with the previous works.
Specifically, in light of the last column for average rank, our method seizes the highest ranking position from SOTA contrastive learning methods as well as self-supervised learning methods, and a significant ranking improvement can be witnessed as opposed to the second place (GraphACL gives the A.R. 4.3).
In particular, LEMON achieves the best performance on all eight benchmarks, and outperforms the best data augmentation model (i.e., GraphACL) with a \textbf{1.87\%} accuracy gain.~\footnote{We do not compare with the results of HGCL because there are several data missing in such work.}
Thus, we can conclude that LEMON captures the molecular semantic information well in the absence of well-designed domain knowledge, and the line graph provides an excellent contrastive view without altering the molecular semantics.

\subsection{Ablation Study}
Here, we delve deeper into the performance influence of each proposed component. First, we analyze the performance boosting from the introduction of the line graph and edge attribute fusion without ZINC15 pre-training. As for the two local contrastive losses, we present the test results of various combinations from these parts following the transfer learning settings. 
Instead of optimizing for SOTA performance, here, we simply set the two hyper-parameters $\alpha$ and $\beta$ to 1 to highlight the inherent advantages of our proposed model components.
The detailed discussions are as follows.

\begin{figure}[!t]
  \centering
  \includegraphics[width=0.8\linewidth]{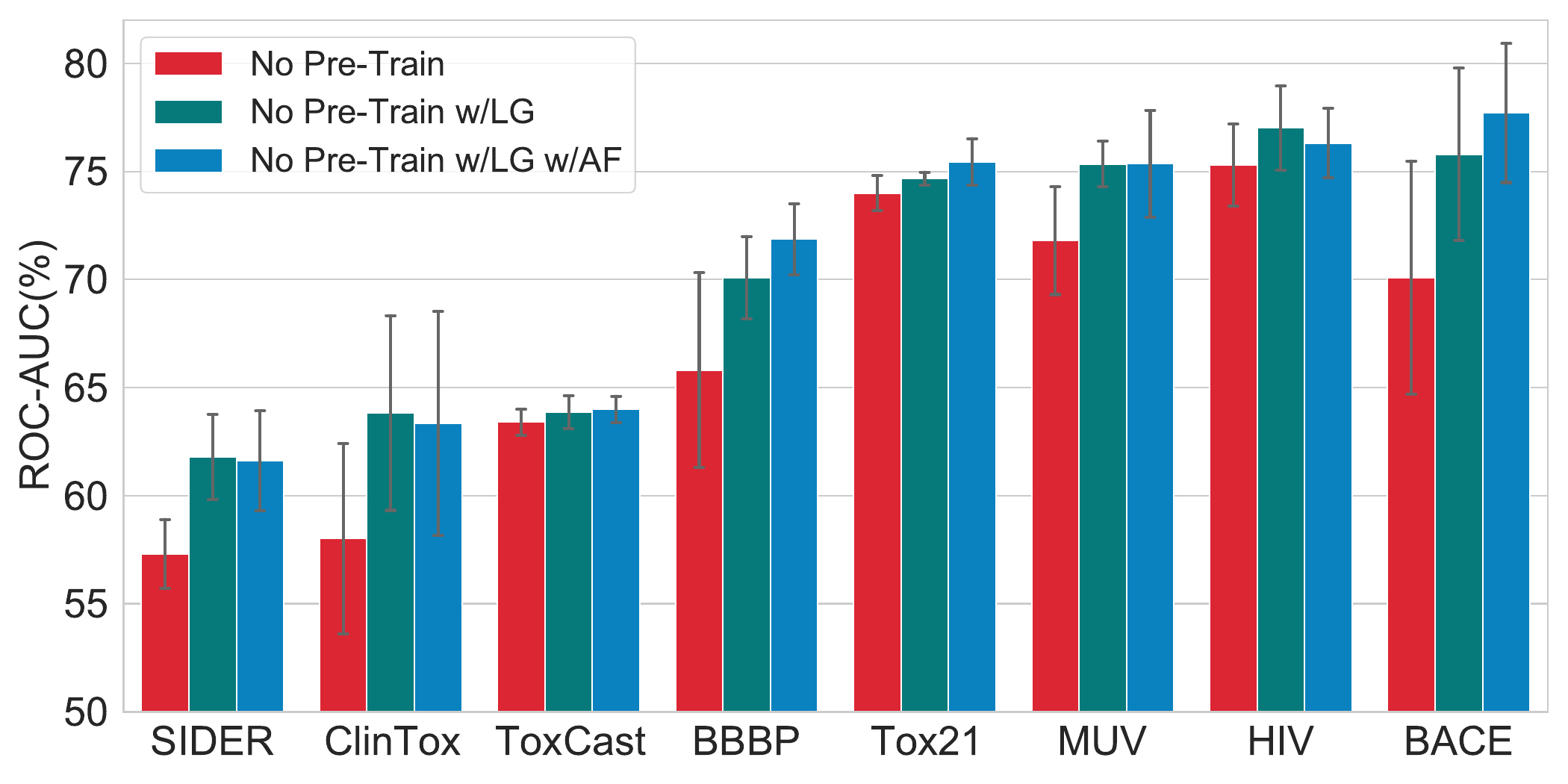}
  \caption{Average test ROC-AUC (\%) gain within `No Pre-Train' from the line graph (w/LG) and edge attribute fusion (w/AF) across all datasets.} 
  \label{fig:no-pretrain-ablation}
\end{figure}

\noindent \textbf{The effectiveness of line graph.}
In Figure~\ref{fig:no-pretrain-ablation}, we analyze the effect of the line graph. In comparison with the red bar (i.e., `No Pre-Train') that denotes the results from random initialization, introducing the line graph (i.e., `No Pre-Train w/LG') shows an overall superior performance, which empirically suggests that the semantics underlying the edges can be better captured by the line graph.

\begin{table}[!t]
\centering
\caption{Average test ROC-AUC (\%) of LEMON with different components. Avg. shows the average ROC-AUC over all datasets. A.R. denotes the average rank. $\alpha$ and $\beta$ are fixed at 1 to highlight the inherent advantages of each component.}
\label{tab:loss_ablation}
\resizebox{0.8\textwidth}{!}{%
\begin{tabular}{lcccccc}
\hline \hline
\# & 1 & 2 & 3 & 4 & 5 & 6 \\ \hline 
LG & \checkmark & \checkmark & \checkmark & \checkmark & \checkmark & \checkmark \\
AF &  &  &  &  & \checkmark & \checkmark \\ \hline
$\mathcal{L}_{G}$ & \checkmark & \checkmark & \checkmark & \checkmark & \checkmark & \checkmark \\
$\mathcal{L}_{IntraC}$ &  & \checkmark &  & \checkmark &  & \checkmark \\
$\mathcal{L}_{InterC}$ &  &  & \checkmark & \checkmark &  & \checkmark \\ \hline \hline
BBBP & 69.97$\pm$2.62 & 70.61$\pm$1.21 & 69.87$\pm$1.27 & 70.94$\pm$0.68 & 70.01$\pm$0.45 & 70.99$\pm$1.05 \\
Tox21 & 75.60$\pm$0.50 & 75.85$\pm$0.49 & 74.97$\pm$0.52 & 75.74$\pm$0.63 & 75.57$\pm$0.56 & 76.95$\pm$0.43 \\
ToxCast & 62.70$\pm$0.55 & 63.83$\pm$0.74 & 64.01$\pm$0.68 & 63.50$\pm$0.65 & 63.59$\pm$0.60 & 64.71$\pm$0.72 \\
SIDER & 59.32$\pm$1.26 & 59.33$\pm$0.51 & 60.49$\pm$1.32 & 60.47$\pm$1.08 & 61.15$\pm$0.49 & 63.37$\pm$0.56 \\
ClinTox & 67.79$\pm$2.52 & 75.82$\pm$2.35 & 76.44$\pm$4.32 & 76.48$\pm$4.82 & 74.95$\pm$3.27 & 77.59$\pm$1.54 \\
MUV & 75.70$\pm$1.75 & 76.37$\pm$1.81 & 75.52$\pm$1.45 & 76.12$\pm$1.93 & 75.97$\pm$1.88 & 77.70$\pm$3.00 \\
HIV & 75.72$\pm$0.84 & 77.32$\pm$0.95 & 77.40$\pm$1.69 & 77.49$\pm$0.79 & 76.51$\pm$1.47 & 78.69$\pm$1.10 \\
BACE & 82.50$\pm$0.73 & 80.79$\pm$1.41 & 84.00$\pm$0.66 & 84.23$\pm$0.87 & 83.67$\pm$1.49 & 84.68$\pm$0.73 \\ \hline
Avg. & 71.16 & 72.49 & 72.84 & 73.12 & 72.68 & 74.33 \\
A.R. & 5.4 & 3.6 & 4.0 & 2.9 & 4.1 & 1.0 \\ \hline \hline
\end{tabular}%
}
\end{table}

\noindent \textbf{The effectiveness of edge attribute fusion.}
Based on the performance boosting of the line graph, we further present the results with edge attribute fusion in Figure~\ref{fig:no-pretrain-ablation} (i.e., `No Pre-Train w/LG w/AF'). 
It is worth noting that there are no results for `No Pre-Train w/AF' because the edge attribution fusion requires the one-to-one correspondence between the vertices of line graph and the edges of original graph.
Besides the general promotion compared with the results of random initialization, the edge attribute fusion also brings five out of eight better results in contrast to the solo line graph. 
Furthermore, following the setting of transfer learning, the performance differences between the first and fifth columns as well as the fourth and sixth columns in Table~\ref{tab:loss_ablation} also validate the effectiveness of edge attribute fusion.
Thus, we may conclude that edge attribute fusion can alleviate information inconsistency and enhance molecular representation learning.

\noindent \textbf{The effectiveness of intra-local contrastive loss.}
The test results under the supervision of the proposed losses are shown in Table~\ref{tab:loss_ablation}. To achieve a comprehensive comparison, we first give a baseline only pre-trained with the NT-Xent loss (see the first column). The effectiveness of the proposed intra-local contrastive loss is confirmed by the performance differences between the second and first columns as well as the fourth and third columns, in which the only experimental setup difference is the $\mathcal{L}_{IntraC}$. Specifically, at least six out of eight better results are obtained via deploying this contrastive loss, which informs us of its effectiveness in over-smoothing addressing.

\noindent \textbf{The effectiveness of inter-local contrastive loss.}
Analogously, when comparing the results of the first and third columns as well as the second and fourth columns in Table~\ref{tab:loss_ablation}, we can observe that six and five datasets achieve performance exaltation, respectively. This metric promotion suggests the effectiveness of the proposed local loss in addressing hard negative samples and improving model performance.
Finally, despite several failures within these ablation studies, the last column that simultaneously adopts all proposed components performs best; thus, the proposed parts of LEMON are complementary to each other in molecular semantic exploration.

\begin{table}[!ht]
\centering
\caption{Average test ROC-AUC (\%) of LEMON with different pre-training datasets. Avg. shows the average ROC-AUC over all datasets.}
\label{tab:abla-pretrain-datasets}
\resizebox{0.8\textwidth}{!}{%
\begin{tabular}{l|ccc}
\hline \hline
Dataset & GEOM-Drugs (304K) & ChEMBL (456K) & ZINC15 (2M) \\  \hline  \hline
BBBP    & 71.19$\pm$0.62 & 72.38$\pm$0.74 & \textbf{73.71$\pm$1.08} \\
Tox21   & 76.75$\pm$0.36 & 76.32$\pm$0.58 & \textbf{77.45$\pm$0.57} \\
ToxCast & 64.99$\pm$0.38 & 64.85$\pm$0.66 & \textbf{65.07$\pm$0.52} \\
SIDER   & \textbf{64.80$\pm$0.48} & 63.38$\pm$0.75 & 64.27$\pm$0.88 \\
ClinTox & 77.06$\pm$0.49 & 79.65$\pm$1.59 & \textbf{85.88$\pm$3.16} \\
MUV     & 76.51$\pm$3.21 & 77.51$\pm$2.71 & \textbf{79.42$\pm$4.26} \\
HIV     & 78.80$\pm$1.61 & 78.64$\pm$1.33 & \textbf{79.34$\pm$1.08} \\
BACE    & 82.31$\pm$1.26 & 83.19$\pm$1.14 & \textbf{87.78$\pm$1.35} \\  \hline
Avg.    & 74.05          & 74.49          & \textbf{76.62}  \\  \hline \hline    
\end{tabular}%
}
\end{table}

\noindent \textbf{Other pre-training datasets.} Besides the superior performance of LEMON pre-trained with 2 million molecules from ZINC15, here, we further validate the effectiveness of our method with other pre-training datasets. Specifically, we utilize the GEOM-Drugs dataset~\cite{axelrod2022geom}, comprising 304k molecules, and the ChEMBL dataset~\cite{gaulton2012chembl}, consisting of 456k molecules. As demonstrated in Table~\ref{tab:abla-pretrain-datasets}, there is a general trend indicating that models pre-trained with larger datasets tend to perform better. However, an exception is observed in the case of SIDER pre-trained with the GEOM-Drugs dataset, which outperforms those pre-trained with the other two datasets. This phenomenon may be attributed to the alignment of molecular features between the SIDER and GEOM-Drugs datasets.

\begin{figure}[!ht]
\centering
\includegraphics[width=0.8\linewidth]{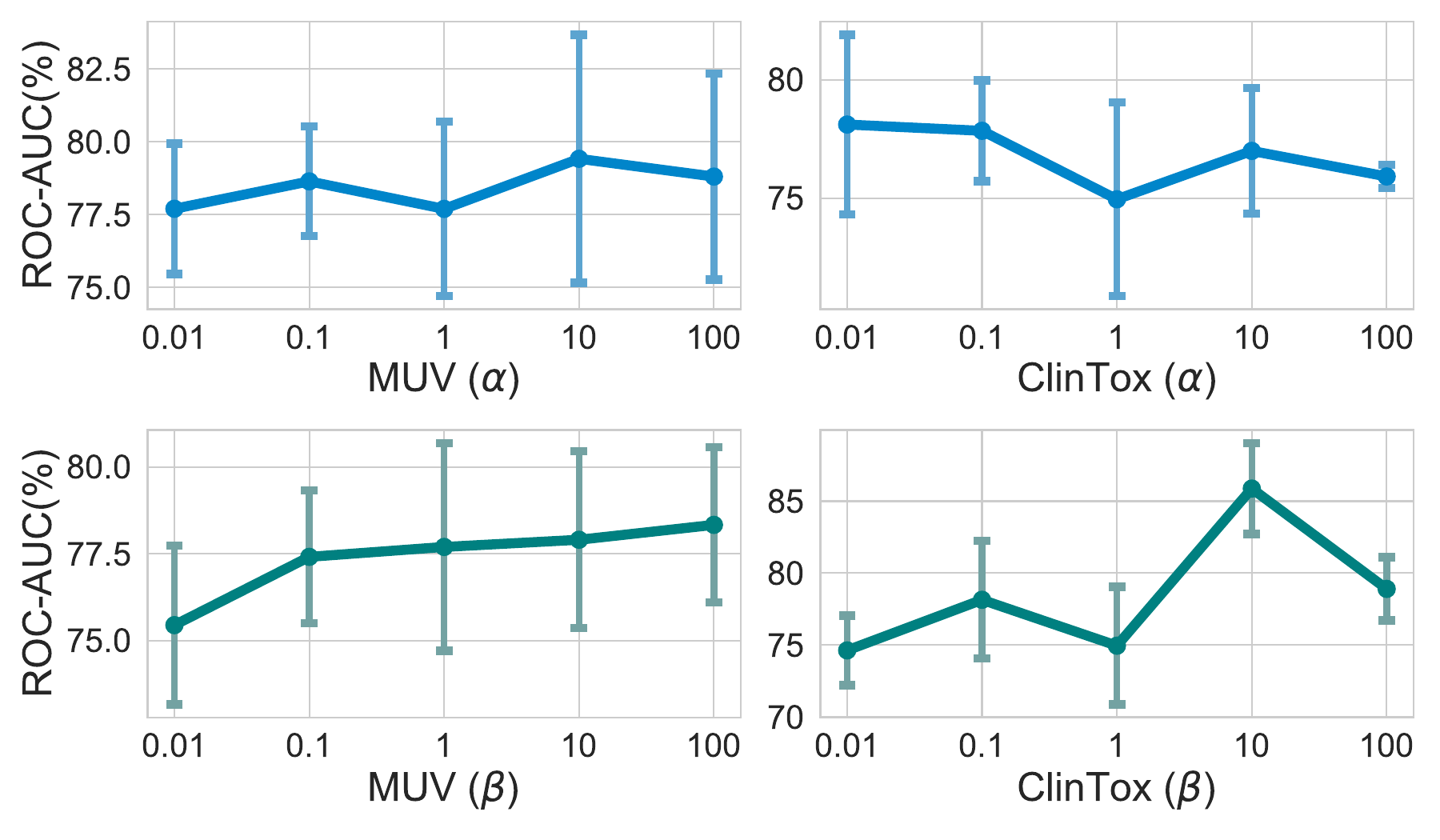}
\caption{Sensitivity \textit{w.r.t.} hyper-parameter $\alpha$ and $\beta$.}
\label{fig:hyper-sensitivity_main}
\end{figure} 

\noindent \textbf{Hyper-parameter sensitivity.}
Here, we show the hyper-parameter sensitivity of LEMON regarding $\alpha$ and $\beta$. In the tunning of $\alpha$, we fix the $\beta$ to 1 and vice versa. The average ROC-AUC scores of downstream tasks are shown in Figure~\ref{fig:hyper-sensitivity_main}.
As can be seen, different downstream tasks prefer different loss controls. 
Specifically, ClinTox prefers small $\alpha$ but large $\beta$, while MUV would like large $\beta$ and is insensitive to $\alpha$, which suggests that ClinTox and MUV suffer more from the over-smoothing.

\begin{figure}[!ht]
  \centering
  \begin{subfigure}{.48\textwidth}
    \centering
    \includegraphics[width=\linewidth]{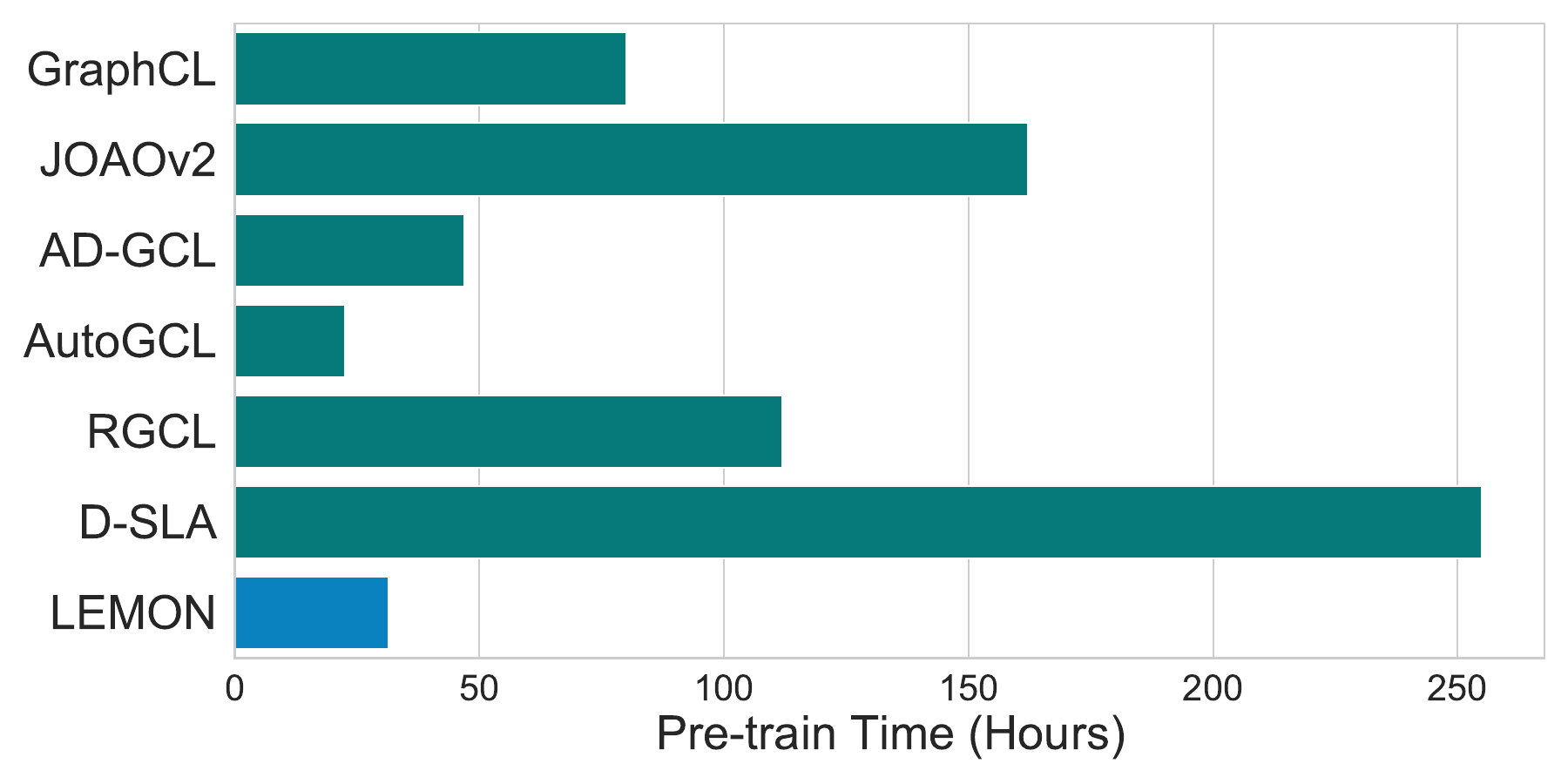}
    \caption{Comparison between baselines and LEMON.}
    \label{fig:time_comparison_baselines}
  \end{subfigure}
  \begin{subfigure}{.48\textwidth}
    \centering
    \includegraphics[width=\linewidth]{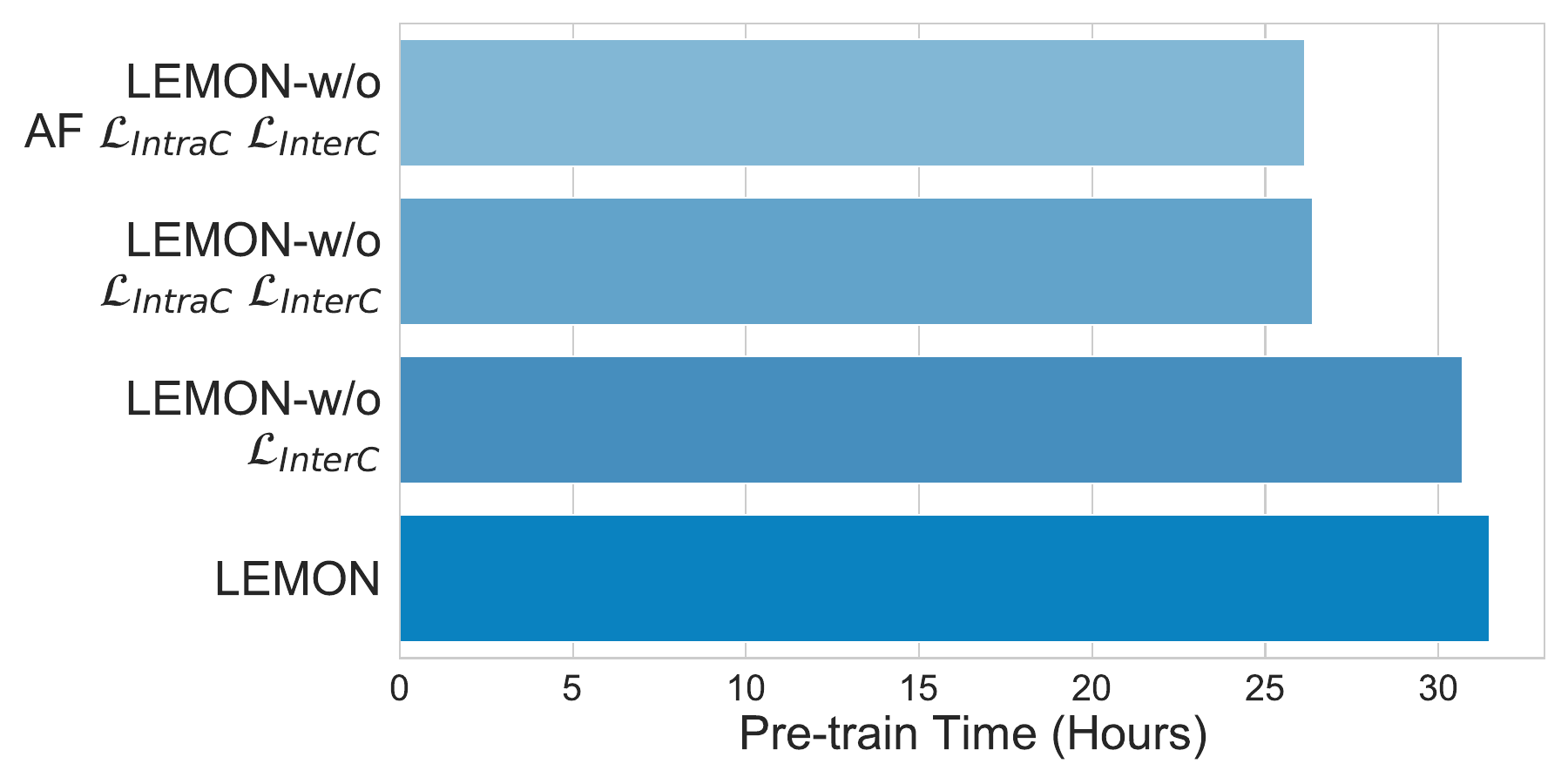}
    \caption{Comparison among parts of LEMON.}
    \label{fig:time_comparison_parts}
  \end{subfigure} 
  \caption{\textbf{Pre-training time comparison.} The time required by LEMON is much less than the time needed by baselines except AutoGCL.} 
  \label{fig:time_comparison}
\end{figure}

\noindent \textbf{Model efficiency.}
Besides the superior performance, we further present the efficiency of LEMON by comparing the pre-training time on 2 million molecular graphs of ZINC15 with the baselines.
In addition, the actual runtime of LEMON through introducing each item is also given, including LEMON with only line graph (i.e., LEMON w/o AF $\mathcal{L}_{IntraC}$ $\mathcal{L}_{InterC}$); LEMON with line graph and edge attribute fusion (i.e., LEMON w/o $\mathcal{L}_{IntraC}$ $\mathcal{L}_{InterC}$); LEMON with line graph, edge attribute fusion and intra-local contrastive loss (i.e., LEMON w/o $\mathcal{L}_{InterC}$).
The pre-training times of the compared baselines and LEMON are evaluated within the same experimental environment (Tesla V100 GPU and Intel(R) Xeon(R) Silver 4214 CPU). Note that, we set the \textit{\color{gray}num\_workers} in \textit{\color{gray}dataloader} to \textit{\color{gray}default} for a fair comparison. 

Figure~\ref{fig:time_comparison} shows the pre-training time required for 2 million molecular graphs from ZINC15 pre-training with 100 epochs. As shown in Figure~\ref{fig:time_comparison_baselines}, because the contrastive views of LEMON are static, the time required by LEMON is much less than the time needed by baselines except AutoGCL, which reveals that LEMON has not only superior performance but also excellent efficiency. In particular, LP-Info requires almost 16 hours for one epoch pre-training, that is about 1,600 hours for 100 epochs, thus we do not report its pre-training time in Figure~\ref{fig:time_comparison_baselines}.

Furthermore, as shown in Figure~\ref{fig:time_comparison_parts}, comparing to the model only with line graph (i.e., LEMON w/o AF $\mathcal{L}_{IntraC}$ $\mathcal{L}_{InterC}$), the additional time from the introduction of edge attribute fusion is nearly negligible. The biggest time consumption gap comes from the proposal of two local contrastive losses, but they still only occupy 16.26\% of the total time, which is a small price in contrast to the accompanying performance boosting.

\section{Conclusions}
In this work, we try to figure out a decent view for molecular GCL that can maintain the integrity of molecular semantic information and is friendly to researchers without profound domain knowledge.
Inspired by the line graph, we propose a method, called LEMON, to tackle our expectations. Due to the different pace of message passing in the original graph and the corresponding line graph, we further present three crucial components to address the concomitant issues and enhance molecular graph representation learning with hard negative samples. 
Under the setting of transfer learning, we empirically present the superior performance of LEMON over SOTA works.

Despite the presented superior performance, the current design of LEMON may be most effective for sparse graphs. As discussed in the section on line graph transformation, a vertex with $e$ edges in the original graph will generate $e\times(e-1)/2$ edges in the corresponding line graph. This could lead to severe runtime complexity when the original graphs are dense, such as in social networks. Therefore, adapting LEMON for dense graphs presents a promising direction for future research.
Furthermore, the existing evaluation protocols for molecular GCL are primarily focused on assessing model performance on molecular property classification tasks. However, there are other tasks related to molecules or sparse graphs that could also be explored, such as molecular regression tasks. These areas could offer valuable opportunities for future research and potential application domains for LEMON.

\section*{Acknowledgements}
This work has been supported by the Guangxi Science and Technology Major Project, China (No. AA22067070), NSFC (Grant No. 61932002), and CCSE project (CCSE-2024ZX-09).

\bibliographystyle{elsarticle-num-names}
\biboptions{compress}
\bibliography{myref}

\clearpage
\appendix

\setcounter{table}{0}
\setcounter{figure}{0}

\section{Unsupervised Learning}
\begin{table}[!ht]
\centering
\caption{Summary statistics of ubiquitous benchmarks from TUDataset.}
\label{tab:data_stat_tu}
\resizebox{\textwidth}{!}{%
\begin{tabular}{l|ccccc}
\hline 
Dataset & \#Graphs & \#Classes & Avg.Nodes & Avg.Degree  & LG Avg.Degree \\ \hline 
\multicolumn{6}{c}{Social Networks} \\ \hline
COLLAB & 5,000 & 3 & 74.49 & 4914.43 & 786967.36 \\
REDDIT-BINARY & 2,000 & 2 & 429.63 & 995.50 & 184826.67 \\
REDDIT-MULTI-5K & 4,999 & 5 & 508.52 & 1189.74 & 81066.29 \\
IMDB-BINARY & 1,000 & 2 & 19.77 & 193.06 & 2782.11 \\
IMDB-MULTI & 1,500 & 3 & 13.00 & 131.87 & 2037.64 \\
GITHUB & 12,725 & 2 & 113.79 & 469.27 & 19574.33 \\ \hline
\multicolumn{6}{c}{Bioinformatics} \\ \hline
\rowcolor{gray!10} MUTAG & 188 & 2 & 17.93 & 39.58 & 57.74 \\
\rowcolor{gray!10} NCI1 & 4,110 & 2 & 29.87 & 64.60 & 93.21 \\
\rowcolor{gray!10} PROTEINS & 1,113 & 2 & 39.06 & 145.63 & 448.99 \\
DD & 1,178 & 2 & 284.32 & 1431.31 & 6581.59 \\ \hline
\end{tabular}%
}
\end{table}

\begin{table}[!ht]
\centering
\caption{Average accuracies (\%)$\pm$Std. of compared methods via unsupervised learning. }
\label{tab:unsupervised}
\begin{tabular}{l|ccc}
\hline
          & NCI1           & PROTEINS       & MUTAG           \\ \hline
GL        & -              & -              & 81.66$\pm$2.11  \\
WL        & 80.01$\pm$0.50 & 72.92$\pm$0.56 & 80.72$\pm$3.00  \\
DGK       & 80.31$\pm$0.46 & 73.30$\pm$0.82 & 87.44$\pm$2.72  \\ \hline
node2vec  & 54.89$\pm$1.61 & 57.49$\pm$3.57 & 72.63$\pm$10.20 \\
sub2vec   & 52.84$\pm$1.47 & 53.03$\pm$5.55 & 61.05$\pm$15.80 \\
graph2vec & 73.22$\pm$1.81 & 73.30$\pm$2.05 & 83.15$\pm$9.25  \\
MVGRL     & -              & -              & 75.40$\pm$7.80  \\
InfoGraph & 76.20$\pm$1.06 & 74.44$\pm$0.31 & 89.01$\pm$1.13  \\
GraphCL   & 77.87$\pm$0.41 & 74.39$\pm$0.45 & 86.80$\pm$1.34  \\
JOAO      & 78.07$\pm$0.47 & 74.55$\pm$0.41 & 87.35$\pm$1.02  \\
JOAOv2    & 78.36$\pm$0.53 & 74.07$\pm$1.10 & 87.67$\pm$0.79  \\
AD-GCL    & 73.91$\pm$0.77 & 73.28$\pm$0.47 & 88.74$\pm$1.85  \\
RGCL      & 78.14$\pm$1.08 & 75.03$\pm$0.43 & 87.66$\pm$1.01  \\
SimGRACE  & 79.12$\pm$0.44 & 75.35$\pm$0.09 & 89.01$\pm$1.31  \\
SEGA      & 79.00$\pm$0.72 & 76.01$\pm$0.42 & 90.21$\pm$0.66  \\
GCS       & 77.37$\pm$0.30 & 75.02$\pm$0.39 & 90.45$\pm$0.81  \\
HGCL      & -              & 75.5$\pm$0.50  & 90.1$\pm$0.80   \\
MACL      & 78.41$\pm$0.47 & 74.47$\pm$0.85 & 89.04$\pm$0.98  \\
GraphACL  & -              & 75.47$\pm$0.38 & 90.21$\pm$0.94  \\ \hline
LEMON     & 79.76$\pm$0.46 & 76.47$\pm$0.39 & 90.32$\pm$1.14  \\ \hline
\end{tabular}
\end{table}

\noindent \textbf{Datasets.} 
Three sparse datasets are adopted from TUDataset \cite{morris2020tudataset} for unsupervised learning, including NCI1 and MUTAG, and PROTEINS. Table~\ref{tab:data_stat_tu} summarizes the characteristics of the three employed datasets.

\begin{itemize}
  \item NCI1 is a dataset made publicly available by the National Cancer Institute (NCI) and is a subset of balanced datasets containing chemical compounds screened for their ability to suppress or inhibit the growth of a panel of human tumor cell lines; this dataset possesses 37 discrete labels. 
  \item MUTAG has seven kinds of graphs that are derived from 188 mutagenic aromatic and heteroaromatic nitro compounds. 
  \item PROTEINS is a dataset where the nodes are secondary structure elements (SSEs), and there is an edge between two nodes if they are neighbors in the given amino acid sequence or in 3D space. The dataset has 3 discrete labels, representing helixes, sheets or turns. 
\end{itemize}

\noindent \textbf{Configuration.} 
To keep in line with GraphCL~\cite{you2020graph}, the same GNN architectures are employed with their original hyper-parameters under individual experiment settings. Specifically, GIN~\cite{xu2019powerful} with 3 layers is set up in unsupervised representation learning. 
The encoder hidden dimensions are fixed for all layers to keep in line with GraphCL under individual experiment setting. 
Models are trained 20 epochs and tested every 10 epochs. 
Hidden dimension is 32, and batch size is $\in \{32, 128\}$. An Adam optimizer \cite{kingma2015adam} is employed to minimize the contrastive lose and learning rate is $\in\{0.01, 0.001, 0.0001\}$. 

\noindent \textbf{Learning protocol.} 
Following the learning setting in SOTA works, the corresponding learning protocols are adopted for a fair comparison. 
In unsupervised representation learning \cite{sun2020infograph}, all data is used for model pre-training and the learned graph embeddings are then fed into a non-linear SVM classifier to perform classification.
Experiments are performed for 5 times each of which corresponds to a 10-fold evaluation as~\cite{sun2020infograph}, with mean and standard deviation of accuracies (\%) reported.

\noindent \textbf{Compared methods.} 
The baselines in unsupervised learning have three categories. 
The first set is three SOTA kernel-based methods that include GL~\cite{shervashidze2009efficient}, WL~\cite{shervashidze2011weisfeiler}, and DGK~\cite{yanardag2015deep}. The second set is four heuristic self-supervised methods, including node2vec \cite{grover2016node2vec}, sub2vec \cite{adhikari2018sub2vec}, graph2vec \cite{narayanan2017graph2vec}, and InfoGraph \cite{sun2020infograph}. The final category comes from the graph contrastive learning domain, including MVGRL \cite{hassani2020contrastive}, GraphCL \cite{you2020graph}, AD-GCL~\cite{suresh2021adversarial}, JOAO~\cite{you2021graph}, RGCL~\cite{li2022let}, SimGRACE~\cite{xia2022simgrace}, SEGA~\cite{wu2023sega}, GCS~\cite{wei2023boosting}, HGCL~\cite{ju2023unsupervised}, MACL~\cite{huang2023model}, and GraphACL~\cite{luo2023self}.

\noindent \textbf{Results.} The results of LEMON along with SOTA competitors on three benchmarks are shown in Table~\ref{tab:unsupervised}. 
To summarize, out method obtains superior performance compared with the previous works.
In particular, except for MUTAG, LEMON achieves the best performance on two out of three benchmarks.

\section{Semi-Supervised Learning}

\noindent \textbf{Datasets.} 
Two datasets are adopted from TUDataset~\cite{morris2020tudataset} for semi-supervised learning, including NCI1 and PROTEINS.

\noindent \textbf{Configuration.}
ResGCN with 128 hidden units and 5 layers is set up in semi-supervised learning. 
For all datasets we perform experiments with 10\% label rate for 5 times, each of which corresponds to a 10-fold evaluation as~\cite{you2020graph}, with mean and standard deviation of accuracies (\%) reported. For pre-training, learning rate is tuned in $\{0.01, 0.001, 0.0001\}$ and epoch number in $\{20, 40, 60, 80, 100\}$ where grid search is performed. 
For fine-tuning, we following the default setting in \cite{you2020graph}, that is, learning rate is 0.001, hidden dimension is 128, bath size is 128, and the pre-trained models are trained 100 epochs.

\noindent \textbf{Learning protocols.} 
Following the learning setting in SOTA works, the corresponding learning protocols are adopted for a fair comparison. 
In semi-supervised learning \cite{you2020graph}, there exist two learning settings. For datasets with a public training/validation/test split, pre-training is performed only on training dataset, finetuning is conducted with 10\% of the training data, and final evaluation results are from the validation/test sets. 
For datasets without such splits, all samples are employed for pre-training while finetuning and evaluation are performed over 10 folds.

\noindent \textbf{Compared methods.} 
Under the setting of semi-supervised learning, ten baselines are adopted, including the naive GCN without pre-training~\cite{you2020graph}, GAE~\cite{kipf2016variational}, Infomax~\cite{velickovic2019deep}, ContextPred~\cite{hu2020strategies}, GraphCL~\cite{you2020graph}, JOAO~\cite{you2021graph}, AD-GCL~\cite{suresh2021adversarial}, SimGRACE~\cite{xia2022simgrace}, and SEGA~\cite{wu2023sega}.

\begin{table}[!ht]
\centering
\caption{Average accuracies (\%) $\pm$ Std. of compared methods via semi-supervised representation learning with 10\% labels.}
\label{tab:semisupervised}
\begin{tabular}{l|cc}
\hline
             & NCI1           & PROTEINS       \\ \hline
No Pre-Train & 73.72$\pm$0.24 & 70.40$\pm$1.51 \\ \hline
GAE          & 74.36$\pm$0.24 & 70.51$\pm$0.17 \\
Infomax      & 74.86$\pm$0.26 & 72.27$\pm$0.40 \\
ContextPred  & 73.00$\pm$0.30 & 70.23$\pm$0.63 \\
GraphCL      & 74.63$\pm$0.25 & 74.17$\pm$0.34 \\
JOAO         & 74.48$\pm$0.27 & 72.13$\pm$0.92 \\
JOAOv2       & 74.86$\pm$0.39 & 73.31$\pm$0.48 \\
AD-GCL       & 75.18$\pm$0.31 & 73.96$\pm$0.47 \\
SimGRACE     & 74.60$\pm$0.41 & 74.03$\pm$0.51 \\
SEGA         & 75.09$\pm$0.22 & 74.65$\pm$0.54 \\ \hline
LEMON        & 75.82$\pm$0.28 & 74.87$\pm$0.39 \\ \hline
\end{tabular}
\end{table}

\noindent \textbf{Results.} 
The results of LEMON along with SOTA competitors on the two benchmarks are shown in Table~\ref{tab:semisupervised}, in which LEMON suppresses the SOTA view generation works on the two adopted datasets.

\section{Visualization}

\begin{figure}[!ht]
\centering
  \begin{minipage}{\textwidth}    
    \begin{figure}[H]
      \centering
      \begin{subfigure}{.48\linewidth}
        \centering
        \includegraphics[width=\linewidth]{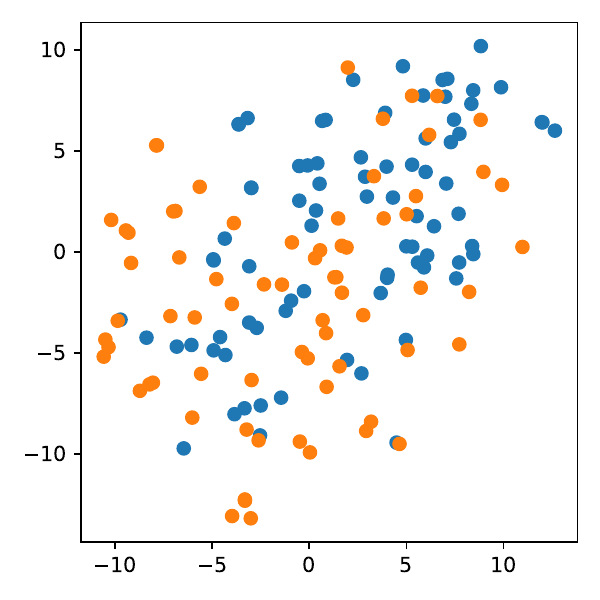}
        \caption{GraphCL.}
        \label{fig:bbbp_graphcl}
      \end{subfigure} 
      \begin{subfigure}{.48\linewidth}
        \centering
        \includegraphics[width=\linewidth]{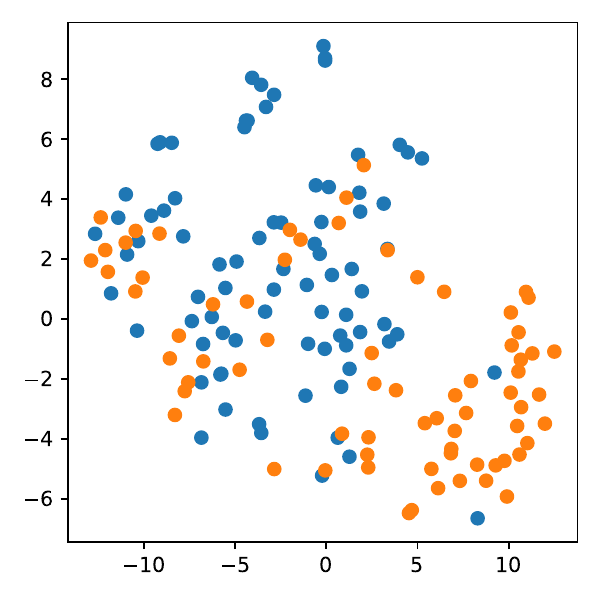}
        \caption{LEMON.}
        \label{fig:bbbp_lgcl}
      \end{subfigure} 
      \caption{\textbf{t-SNE visualization of the graph embedding on BBBP.}}
      \label{fig:bbbp_visual}
    \end{figure}
  \end{minipage}
  \begin{minipage}{\textwidth}    
    \begin{figure}[H]
      \centering
      \begin{subfigure}{.48\linewidth}
        \centering
        \includegraphics[width=\linewidth]{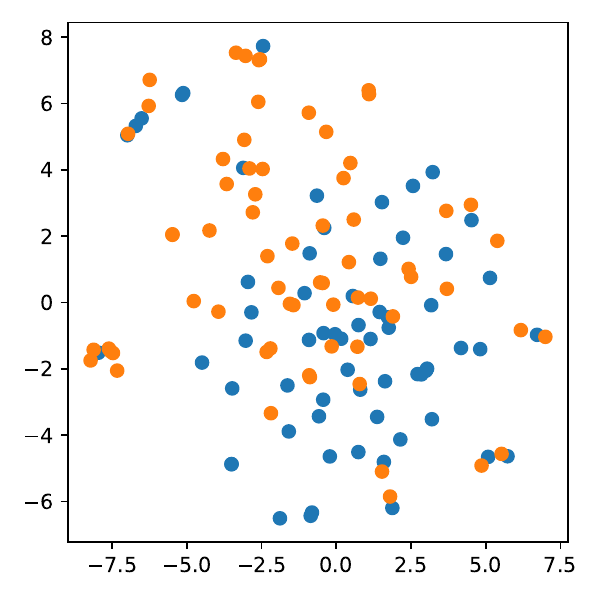}
        \caption{GraphCL.}
        \label{fig:bace_graphcl}
      \end{subfigure} 
      \begin{subfigure}{.48\linewidth}
        \centering
        \includegraphics[width=\linewidth]{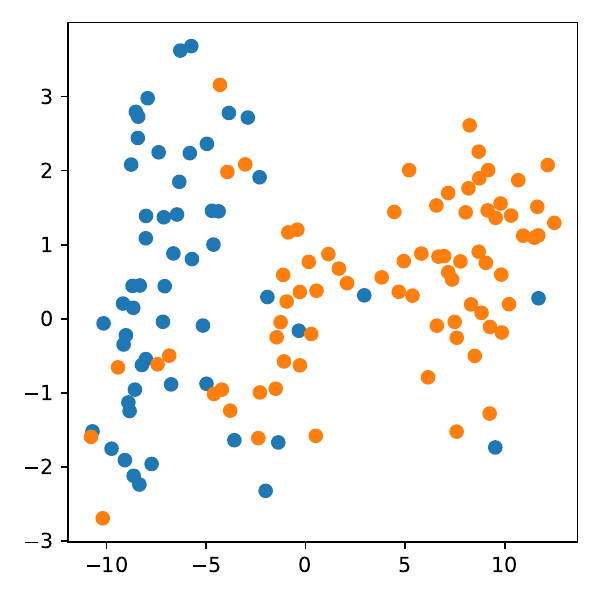}
        \caption{LEMON.}
        \label{fig:bace_lemon}
      \end{subfigure} 
      \caption{\textbf{t-SNE visualization of the graph embedding on BACE.}}
      \label{fig:bace_visual}
    \end{figure}
  \end{minipage}
\end{figure}

To further reveal the superiority of our method in molecular representation learning, we visualize the learned hidden embeddings of GraphCL and LEMON on the BBBP and BACE datasets. The comparative visualizations are shown in Figure~\ref{fig:bbbp_visual} and Figure~\ref{fig:bace_visual}. As can be seen, compared to GraphCL, which is pre-trained through data corruption, the node hidden embeddings learned via our method exhibit a more distinct separation between classes, indicating that our method learns chemically meaningful representations.

Specifically, the enhanced performance on BBBP offers significant advancements in the selection process of compound libraries. This improvement not only reduces the financial burden of experimental testing but also prioritizes compounds with promising permeability rates, ensuring efficient resource allocation. Furthermore, analyzing molecules with diverse permeability rates yields valuable insights into the physicochemical and structural determinants that govern the blood-brain barrier's permeability. Such knowledge is instrumental in driving forward research and comprehension of the blood-brain barrier mechanisms. Concurrently, this understanding is applicable to the investigation of other molecules pertinent to central nervous system disorders, including biomarkers and diagnostic agents, thus enhancing the breadth of neurological research.

\end{document}